\documentclass[sigconf]{acmart}
\usepackage{amsmath,graphicx}
\usepackage{enumitem}
\usepackage{algorithm}
\usepackage{algorithmic}
\usepackage{float}
\usepackage{graphicx}
\usepackage{subcaption}
\usepackage{multirow}

\usepackage{hyperref}

\AtBeginDocument{%
  }

\copyrightyear{2026}
\acmYear{2026}
\setcopyright{cc}
\setcctype{by}
\acmConference[ICMR '26]{International Conference on Multimedia Retrieval}{June 16--19, 2026}{Amsterdam, Netherlands}
\acmBooktitle{International Conference on Multimedia Retrieval (ICMR '26), June 16--19, 2026, Amsterdam, Netherlands}
\acmDOI{10.1145/3805622.3810801}
\acmISBN{979-8-4007-2617-0/2026/06}

\begin{document}

\title{VideoAgent: Personalized Synthesis of Scientific Videos}

\settopmatter{authorsperrow=4}
\author{Xiao Liang}
\authornote{Both authors contributed equally to this research.}
\email{ecoxial2012@outlook.com}
\affiliation{%
  \institution{RobotY, Xidian University}
  \city{Xi'an}
  \country{China}
}

\author{Bangxin Li}
\authornotemark[1]
\email{libangxin@stu.xidian.edu.cn}
\affiliation{%
  \institution{ICTT, Xidian University}
  \city{Xi'an}
  \country{China}
}

\author{Zixuan Chen}
\email{zixuanchen@stu.xidian.edu.cn}
\affiliation{%
  \institution{ICTT, Xidian University}
  \city{Xi'an}
  \country{China}
}

\author{Hanyue Zheng}
\email{zhenghy@stu.xidian.edu.cn}
\affiliation{%
  \institution{ICTT, Xidian University}
  \city{Xi'an}
  \country{China}
}

\author{Zhi Ma}
\authornote{Corresponding authors.}
\email{mazhi@xidian.edu.cn}
\affiliation{%
  \institution{ICTT, Xidian University}
  \city{Xi'an}
  \country{China}
}

\author{Di Wang}
\email{wangdi@xidian.edu.cn}
\affiliation{%
  \institution{RobotY, Xidian University}
  \city{Xi'an}
  \country{China}
}

\author{Cong Tian}
\email{ctian@mail.xidian.edu.cn}
\affiliation{%
  \institution{ICTT, Xidian University}
  \city{Xi'an}
  \country{China}
}

\author{Quan Wang}
\authornotemark[2]
\email{qwang@xidian.edu.cn}
\affiliation{%
  \institution{RobotY, Xidian University}
  \city{Xi'an}
  \country{China}
}

\renewcommand{\shortauthors}{Liang et al.}

\begin{abstract}
The technical complexity of research papers often limits their reach, necessitating more accessible formats like scientific videos to disseminate key insights through engaging narration. However, existing automated methods primarily focus on static posters or slide presentations that remain template-bound and linear. Shifting to audience-adaptive video synthesis requires addressing non-linear narrative orchestration and the joint synchronization of disparate multimodal assets. We introduce \textbf{\textit{VideoAgent}}, a modular framework that redefines scientific video synthesis as an intent-driven planning problem. By decoupling content understanding from multimodal synthesis, \textbf{\textit{VideoAgent}} adaptively interleaves static slides with dynamic animations to match the semantic density of the narration. We further propose \textbf{\textit{SciVidEval}}, a benchmark evaluating multimodal quality and pedagogical utility through automated metrics and human knowledge transfer studies. Extensive experiments demonstrate that VideoAgent effectively conveys complex technical logic with high narrative fidelity and communicative impact.
\end{abstract}



\maketitle

\section{Introduction}
\label{sec:intro}

The technical complexity of research papers often limits their reach beyond expert circles, necessitating more accessible formats for disseminating key insights. Previous efforts to automate this process have primarily focused on static media, such as automated poster generation \cite{pang2025paper2poster, xu2022posterbot, qiang2016learning, saxena2025postersum} and slide presentation creation \cite{SunHWZW21D2S, zheng2025pptagent}. While these paradigms have achieved significant success in multimodal extraction and layout synthesis, they remain largely template-bound and linear, resulting in static renderings that lack expressive narration. In contrast, personalized scientific videos offer a far more engaging medium by illustrating temporal processes and adapting explanations to specific viewer backgrounds \cite{Salzmann2025AND}. By modulating narrative structure and visual emphasis, video can sustain attention and clarify intricate reasoning in ways static formats cannot. However, shifting from the established paradigms of static layout generation to such dynamic, audience-adaptive video synthesis is a non-trivial task. It requires addressing several core challenges: 1) \textbf{Personalized and Dynamic Orchestration.} Unlike template-based methods, personalized scientific video synthesis requires non-linearly restructuring technical documents into intent-driven personalized narrative flows. The core challenge lies in adaptively orchestrating heterogeneous visual forms—such as static slides and dynamic animations—into a coherent story without relying on a rigid, predetermined storyboard. 2) \textbf{Multimodal Content Synchronization.} Beyond narrative planning, achieving a cohesive video requires orchestrating temporal alignment and semantic consistency across disparate assets. The difficulty lies in dynamically synchronizing generated narration with varying visual sources in an automated pipeline, necessitating a unified strategy to resolve temporal dependencies while ensuring visual emphasis matches the spoken explanation.

\begin{figure*}[!t]
    \centering
    \includegraphics[width=\textwidth]{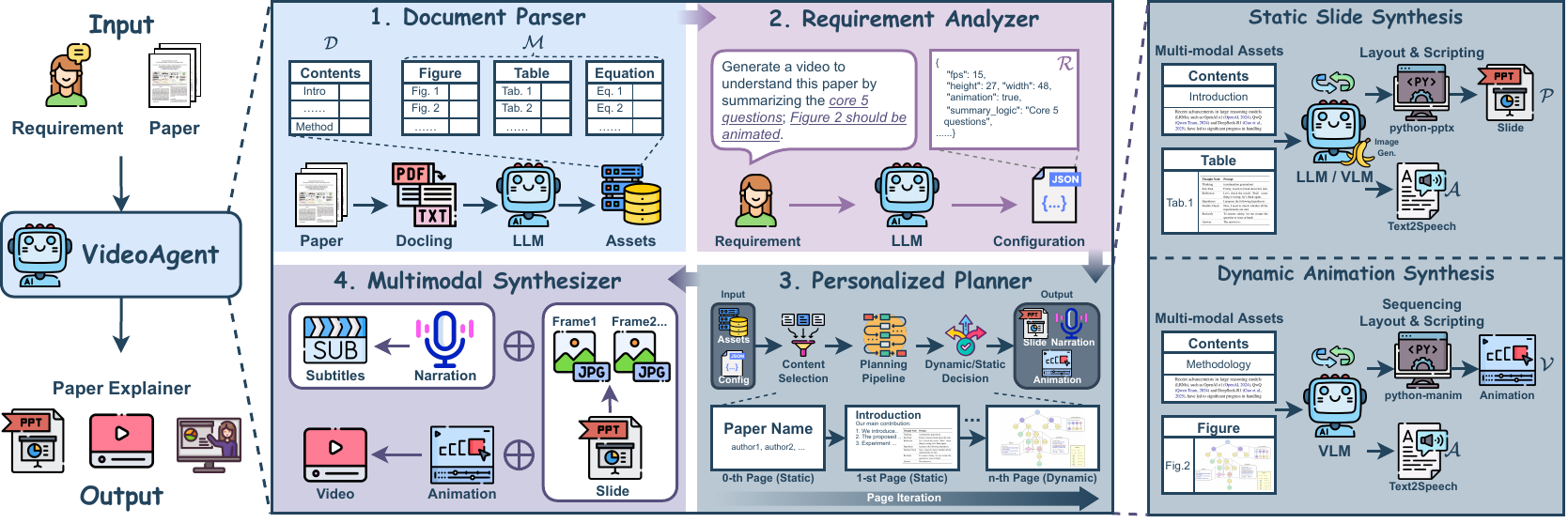}
    \vspace{-0.7cm}
    \caption{Overview of the \textbf{\textit{VideoAgent}} framework.}
    \label{fig:framework}
\end{figure*}

To address these, we introduce \textbf{\textit{VideoAgent}}, a modular, multi-stage framework redefining scientific video synthesis as intent-driven planning. Unlike prior presentation generators constrained by rigid, linear mappings from source documents, \textbf{\textit{VideoAgent}} leverages a staged design to decouple content understanding, narrative planning, and multimodal realization. As illustrated in Figure~\ref{fig:framework}, the framework first decomposes a technical paper into fine-grained assets, which are then dynamically reorganized into a non-linear outline guided by explicit user intent. To achieve seamless synchronization, the system jointly orchestrates the selection and timing of heterogeneous visual forms—interleaving static slides with dynamic animations—to match the semantic density of the narration. By resolving temporal dependencies across disparate modalities in a unified pipeline, \textbf{\textit{VideoAgent}} ensures that the resulting video is not only narratively coherent but also precisely aligned with the underlying technical logic, enabling flexible personalization for diverse audience backgrounds. To evaluate the communicative effectiveness of synthesized videos, we introduce \textbf{\textit{SciVidEval}}, a framework evaluating both multimodal quality and pedagogical utility. We quantify generation fidelity across three core dimensions—narrative, visual, and synchronization—while assessing knowledge acquisition through a Video-Quiz-based human study with graduate-level participants. This allows us to measure how effectively key scientific ideas are conveyed beyond surface-level aesthetics. Our main contributions are threefold:
\vspace{-5pt}
\begin{itemize}[left=0pt]
\item We propose \textbf{\textit{VideoAgent}}, a framework formulating scientific video generation as personalized planning guided by user intent to adaptively orchestrate static slides and dynamic animations.

\item We introduce \textbf{\textit{SciVidEval}}, a benchmark bridging automated metrics and human quizzes to evaluate scientific presentation quality and knowledge transfer.

\item Extensive experiments on real conference paper video pairs demonstrate strong visual quality and effective knowledge transfer, thereby significantly enhancing scientific communication.
\end{itemize}


\vspace{-5pt}
\section{Related Work}

\subsection{Scientific Presentation Synthesis}

Research in automatic scientific presentation synthesis aims to transform scholarly knowledge into visual formats such as posters, slides, or videos. Early studies treated this task as a combination of content extraction and template filling~\cite{qiang2016learning, xu2022posterbot, SunHWZW21D2S, Fu2021DOC2PPTAP}, but these methods rely heavily on predefined templates and handcrafted rules, limiting output diversity and global structural coherence. With the advancement of LLMs, multi-agent frameworks have significantly enhanced generation quality. PosterAgent~\cite{pang2025paper2poster} addresses long-context challenges through hierarchical content summarization, while PPTAgent~\cite{zheng2025pptagent} and PreGenie~\cite{Xu2025PreGenieAA} integrate self-correction modules with visual-aware feedback. For multimodal coordination, PASS~\cite{Aggarwal2025PASSPA} and PresentAgent~\cite{Shi2025PresentAgentMA} explore the alignment of visual and auditory information, and PresentCoach~\cite{Chen2025PresentCoachDP} leverages voice cloning to enhance expressiveness. Despite these strides, most methods assume that presentations should follow the linear logic of the paper. However, \citet{maheshwari-etal-2024-presentations} point out that effective oral presentations are often non-linear. Moreover, existing methods are confined to static outputs, overlooking video's capacity for temporal illustration. Manimator~\cite{P2025Manimator} addresses this gap by using LLMs to generate Manim animations from research papers, yet relies on sampling multiple independent generations for robustness and lacks iterative refinement mechanisms. \textbf{\textit{VideoAgent}} differs in two aspects: 1) a diverse outline generation mechanism for non-linear storytelling, and 2) a planning-first animation synthesis module with self-correction that improves first-pass success without repeated sampling.
\vspace{-5pt}
\subsection{Evaluation Metrics and Benchmarking}

Evaluating scientific presentation quality is inherently challenging because of its multi dimensional nature. Early research primarily relied on general NLP metrics like Rouge \cite{lin-2004-rouge} to measure textual overlap, but these methods ignore semantic consistency and multimodal structural nuances. Recent studies have expanded evaluation into three core dimensions: 1) Content Fidelity, ensuring hallucination-free alignment: D2S \cite{SunHWZW21D2S} employs query-driven retrieval for segment alignment, while SlideTailor \cite{Zeng2025SlideTailorPP} and PPTAgent \cite{zheng2025pptagent} assess content structure and subtopic coverage to ensure cross-modal fidelity. 2) Visual Aesthetics, measuring layout professionalism and styling: SlideGen \cite{Liang2025SlideGenCM} introduces the geometry-aware density score for spatial balance, whereas PosterGen \cite{Zhang2025PosterGenAP} and PosterForest \cite{Choi2025PosterForestHM} optimize layout professionalism through hierarchical design rubrics and tree-based modeling. 3) Functional Utility, verifying information transfer: Paper2Poster \cite{pang2025paper2poster} evaluates knowledge conveyance via PaperQuiz, while SlideBot \cite{Xie2025SlideBotAM} and Auto-Slides\cite{Yang2025AutoSlidesAI} enhance information transfer efficiency by incorporating cognitive science principles and interactive customization. Beyond these static dimensions, video evaluation introduces additional challenges in temporal dynamics and audio-visual synchronization, necessitating automated toolchains for asset extraction and alignment. To address this gap, we propose \textbf{\textit{SciVidEval}}, a comprehensive benchmark for scientific video synthesis. It integrates automated metrics for multimodal quality and synchronization with a Video-Quiz-based human evaluation to rigorously measure knowledge transfer.

\section{VideoAgent}
\label{sec:method}
To address personalized orchestration and multimodal synchronization in scientific video generation, we propose \textit{\textbf{VideoAgent}}, an intent-driven synthesis framework (Figure~\ref{fig:framework}). We detail the modules below: document parsing (\S~\ref{sec:parser}), requirement analysis (\S~\ref{sec:analyzer}), personalized planning (\S~\ref{sec:planner}), and multimodal synthesis (\S~\ref{sec:synthesizer}).

\begin{table*}[htbp]
  \centering
  \caption{Comparison of different personalization methodologies and their generated narrative outlines.}
  \vspace{-0.35cm}
  \footnotesize
  \setlength{\tabcolsep}{3pt}
  \renewcommand{\arraystretch}{1.0}
  \begin{tabular}{@{}p{0.31\textwidth}p{0.46\textwidth}p{0.17\textwidth}@{}}
    \toprule
    \textbf{Methodology} & \textbf{Content Requirements ($\mathcal{R}_{\text{cont}}$)} & \textbf{Slide Outline Example ($O$)} \\
    \midrule

    \parbox[t]{\linewidth}{\raggedright
    \textbf{Default Sequential Summary}\\
    \small \textbf{Explanation:} Follows the original logical flow of the manuscript.}
    &
    \parbox[t]{\linewidth}{\raggedright\small
    You are the author of the paper. Please summarize the core points in sequence, following the logical structure of the manuscript's chapters.}
    &
    \parbox[t]{\linewidth}{\raggedright\small
    Introduction \& Motivation\\
    Methodology Overview\\
    {…}}
    \\
    \midrule

    \parbox[t]{\linewidth}{\raggedright
    \textbf{Knowledge Distillation Paradigm}\\
    \small \textbf{Explanation:} Distills core structures to avoid linear walkthroughs.}
    &
    \parbox[t]{\linewidth}{\raggedright\small
    You are an expert academic planner specializing in the ``Core Five Questions'' methodology. Your task is to analyze a research paper and extract its core structure by identifying five fundamental questions.}
    &
    \parbox[t]{\linewidth}{\raggedright\small
    What is the problem?\\
    Why does it matter?\\
    How do we solve it?{…}\\
    }
    \\
    \midrule

    \parbox[t]{\linewidth}{\raggedright
    \textbf{\mbox{Audience-Adaptive Narrative Modeling}}\\
    \small \textbf{Explanation:} Adapts personas and complexity for specific backgrounds.}
    &
    \parbox[t]{\linewidth}{\raggedright\small
    You are a passionate middle school teacher. Explain each section of this paper to a thirteen-year-old middle school student in simple, easy-to-understand language with analogies.}
    &
    \parbox[t]{\linewidth}{\raggedright\small
    Meet the Team Behind {…}\\
    What is {…} All about?\\
    Why This is Cool?{…}\\
    }
    \\
    \midrule

    \parbox[t]{\linewidth}{\raggedright
    \textbf{Heuristic Insight Discovery}\\
    \small \textbf{Explanation:} Identifies cross-domain connections and practical innovation.}
    &
    \parbox[t]{\linewidth}{\raggedright\small
    You are a Tech Consultant. Your goal is to transform academic research papers into practical, actionable project plans and highlight hidden innovative insights.}
    &
    \parbox[t]{\linewidth}{\raggedright\small
    Step 1: Understanding {…}\\
    Step 2: Designing {…}\\
    {…}}
    \\
    \midrule

    \parbox[t]{\linewidth}{\raggedright
    \textbf{Dialectical \& Critical Synthesis}\\
    \small \textbf{Explanation:} Simulates multi-persona debates to present technical depth.}
    &
    \parbox[t]{\linewidth}{\raggedright\small
    You are a dual-persona explainer. Simultaneously play the ``Host'' and the ``Skeptic'' to deep-dive into the paper's core contributions and potential limitations.}
    &
    \parbox[t]{\linewidth}{\raggedright\small
    This sounds like {…}\\
    Isn't {…} too complex?\\
    {…}}
    \\
    \bottomrule
  \end{tabular}
  \label{tab:outline_example}
\end{table*}

\subsection{Document Parser}
\label{sec:parser}

Scientific papers are typically lengthy and possess complex internal hierarchies. Performing effective orchestration directly on raw text is difficult because it lacks the semantic boundaries and asset associations required for narrative reasoning. To support reliable global planning, the Document Parser identifies the paper's inherent structure and transforms the unstructured PDF into two complementary artifacts: a refined textual representation $\mathcal{D}$ and a visual asset library $\mathcal{M}$. It initially employs Docling~\cite{livathinos2025docling} to parse the PDF into structured \texttt{Markdown} format. For papers with highly complex layouts---such as multi-column text interleaved with dense tables and figures---where structural continuity might be compromised, the pipeline leverages Marker~\cite{paruchuri2023marker} to perform layout-aware reconstruction, ensuring the completeness of the extracted narrative elements. The extracted text is reorganized into a structured \texttt{JSON} representation $\mathcal{D}$ that ensures structural completeness and format validity. In parallel, all visual elements are extracted: equations are preserved as high-resolution image crops, while figures and tables are indexed with captions and layout metadata to form the visual asset library $\mathcal{M}$. By decoupling semantic content from reusable visual assets, the Document Parser enables downstream modules to reason about narrative structure over well-defined section-level content blocks and indexed visual assets.

\subsection{Requirement Analyzer}
\label{sec:analyzer}

Personalized scientific video generation requires transforming underspecified, high-level user intent into explicit constraints that can guide narrative planning and multimodal synthesis. This task is particularly challenging because natural language requests such as target audience, emphasis preferences, or presentation style are inherently underspecified and cannot be reliably enforced by fixed templates. The Requirement Analyzer addresses this gap by formalizing user intent into a structured configuration $\mathcal{R}$~\cite{ma2025bridging}, which serves as the decision basis for all subsequent orchestration stages. Specifically, the analyzer decomposes user intent into three complementary dimensions, each governing a distinct aspect of the generation process:


\begin{itemize}[left=0pt]
    \item \textbf{Technical Specifications ($\mathcal{R}_{\text{spec}}$).}
    This dimension defines low-level presentation parameters, including video resolution, frame rate, typography, and color schemes.
    These constraints ensure visual consistency across different output formats and display environments, while remaining independent of narrative logic.

    \item \textbf{Functional Requirements ($\mathcal{R}_{\text{func}}$).}
    Functional requirements specify which generation capabilities are enabled in the pipeline, such as dynamic animation synthesis, synchronized voiceover narration, or multimodal image generation.
    By explicitly exposing these options, $\mathcal{R}_{\text{func}}$ acts as a control gate that constrains downstream planning decisions, determining which visual realization strategies are admissible.
    
    \item \textbf{Content Requirements ($\mathcal{R}_{\text{cont}}$).}
    This dimension governs how the source paper is reorganized into a narrative structure.
    Rather than enforcing a fixed, linear walkthrough of the manuscript, VideoAgent supports multiple knowledge distillation paradigms that reshape content according to audience background and communicative goals.
    As shown in Table~\ref{tab:outline_example}, these paradigms enable non-linear, storytelling-oriented outlines and treat narrative structure as an explicit planning variable.
\end{itemize}

\vspace{-6pt}

The configuration $\mathcal{R}$ is represented in a unified \texttt{JSON} format, serving as the authoritative specification for the Personalized Planner. By translating ambiguous user intent into explicit narrative and modality constraints, the Requirement Analyzer enables structured orchestration and facilitates dynamic, personalized video synthesis.


\vspace{-10pt}

\subsection{Personalized Planner}
\label{sec:planner}
After explicitly formalizing the refined textual representation $\mathcal{D}$, the visual asset library $\mathcal{M}$, and user intent $\mathcal{R}$, the core challenge reduces to a planning problem: constructing a coherent, personalized multimodal presentation path. Considering the intricate interdependencies among narrative structure, visual forms, and generation modes, we introduce the Personalized Planner to integrate these artifacts into a unified, executable planning space. To perform the necessary reasoning, the planner invokes a suite of LLM-driven operators $\mathcal{F}$, each acting as a specialized agent for a distinct coordination task. Operationally, the planner proceeds in two sequential phases: narrative planning followed by execution-and-verification, with $\mathcal{R}_{\text{func}}$ serving as the control gate. In the narrative planning phase, the planner generates a full narrative outline $\mathcal{O} = \{o_i\}_{i=1}^n$ and concurrently allocates multimodal assets globally $\mathcal{M}_{alloc} = \{m_i\}_{i=1}^n$:
\vspace{-5pt}
\begin{equation}
    \mathcal{O}, \mathcal{M}_{alloc} = \mathcal{F}_{\text{outline}}(\mathcal{D}, \mathcal{M}, \mathcal{R}_{\text{cont}}).
\end{equation} Based on the required presentation modality and the control signals in $\mathcal{R}_{\text{func}}$, the synthesis process is bifurcated into two specialized pathways—\textbf{Static Slide Synthesis} and \textbf{Dynamic Animation Synthesis}—coordinated through a unified execution pipeline as detailed in Algorithm~\ref{alg:synthesis}.

\begin{algorithm}[h]
\caption{Personalized Planner}
\label{alg:synthesis}
\begin{algorithmic}[1]
\REQUIRE Narrative Outline $\mathcal{O}$, Allocated Assets $\mathcal{M}_{alloc}$, Source Content $\mathcal{D}$, Configuration $\mathcal{R}$
\ENSURE Presentation $\mathcal{P}$, Audio Set $\mathcal{A}$, Animation Clips $\mathcal{V}$
\FOR{each narrative node $(o_i, m_i) \in (\mathcal{O}, \mathcal{M}_{alloc})$}
    \IF{animation enabled in $\mathcal{R}_{\text{func}}$}
        \STATE $o_i \gets \mathcal{F}_{\text{refine}}(o_i, m_i, \mathcal{D})$
        \STATE $c_{anim,i} \gets \mathcal{F}_{\text{anim}}(o_i, \mathcal{R}_{\text{spec}})$
        \STATE $c_i \gets \mathcal{F}_{\text{layout}}(\emptyset)$
    \ELSE
        \STATE $v_i \gets \text{if visual abstract enabled } \mathcal{F}_{\text{vgen}}(o_i, m_i, \mathcal{R}_{\text{spec}}) \text{ else } \emptyset$
        \STATE $c_i \gets \mathcal{F}_{\text{layout}}(o_i, v_i, m_i, \mathcal{R}_{\text{spec}})$
    \ENDIF
    \STATE $a_i \gets \text{if narration enabled } \mathcal{F}_{\text{narr}}(o_i, \mathcal{R}_{\text{spec}}) \text{ else } \emptyset$
\ENDFOR
\REPEAT
    \STATE $\mathcal{C}_{slide} \gets \text{Concat}(\{c_i\}_{i=1}^n)$
    \STATE $(\textit{status}_s, \mathcal{P}) \gets \textsc{Exec}(\mathcal{C}_{slide})$
    \STATE $(\textit{status}_d, \mathcal{V}) \gets \{\textsc{Exec}(c_{anim,i}) \mid \text{animation enabled}\}$
    \STATE $\mathcal{A} \gets \{a_i\}_{i=1}^n$
    \STATE $\textit{valid} \gets \mathcal{F}_{\text{verify}}(\mathcal{P}, \mathcal{V})$
\UNTIL{$\textit{status}=\text{Success}$ AND $\textit{valid}=\text{True}$}
\RETURN $\mathcal{P}, \mathcal{A}, \mathcal{V}$
\end{algorithmic}
\end{algorithm}

\begin{itemize}[left=0pt]
    \item \textbf{Static Slide Synthesis.} For each narrative node $o_i$ and its allocated asset $m_i$, the planner generates implementation code $c_i$ and an optional narration script $a_i$. When visual abstraction is enabled in $\mathcal{R}_{\text{func}}$, the planner invokes a multimodal synthesizer (e.g., Nano Banana~\cite{google2025nanobanana}) to generate a thematic visual abstract $v_i = \mathcal{F}_{\text{vgen}}(o_i, m_i, \mathcal{R}_{\text{spec}})$. The implementation code $c_i$ is then determined based on the execution mode defined in $\mathcal{R}_{\text{func}}$:

    \begin{equation}
        c_i = 
        \begin{cases} 
            \mathcal{F}_{\text{layout}}(\emptyset) & \text{if animation enabled in } \mathcal{R}_{\text{func}} \\
            \mathcal{F}_{\text{layout}}(o_i, v_i, m_i, \mathcal{R}_{\text{spec}}) & \text{otherwise}
        \end{cases}
    \end{equation}
    
    In the static layout case, the operator $\mathcal{F}_{\text{layout}}$ orchestrates the spatial arrangement based on the availability of visual components. Specifically, if the visual abstract $v_i$ exists, a global background or central layout is applied. If only the asset $m_i$ exists, the operator automatically selects between a right-side or bottom-side placement depending on the asset's aspect ratio. In cases where no visual assets are present, the operator utilizes the narrative text $o_i$ to generate bulleted summaries that fill the slide area. For nodes designated for procedural animation, $\mathcal{F}_{\text{layout}}$ generates a blank placeholder to reserve slots for subsequent Manim integration. Furthermore, if narration is enabled in $\mathcal{R}_{\text{func}}$, the planner generates a corresponding narration script $a_i = \mathcal{F}_{\text{narr}}(o_i, \mathcal{R}_{\text{spec}})$. To finalize the static presentation, the planner concatenates all generated code snippets into a unified \texttt{python} script:
    \begin{equation}
        \mathcal{C}_{slide} = \text{Concat}(\{c_i\}_{i=1}^n)
    \end{equation}
    This script is then processed via $\textsc{Exec}(\mathcal{C}_{slide})$ using \texttt{python-pptx} to generate the final \texttt{.pptx} document $\mathcal{P}$.

\vspace{2pt}
        \item \textbf{Dynamic Animation Synthesis.} When animation is enabled in $\mathcal{R}_{\text{func}}$ for narrative segments requiring in-depth explanation of procedural content such as workflows or algorithms, the planner leverages \texttt{Manim} to generate procedural animations. Specifically, the operator $\mathcal{F}_{\text{refine}}$ refines the coarse narrative node $o_i$ into a detailed storyboard by extracting technical details from the source textual representation $\mathcal{D}$ to ensure scientific grounding:
    \begin{equation}
        o_i \gets \mathcal{F}_{\text{refine}}(o_i, m_i, \mathcal{D})
    \end{equation}
    The resulting storyboard incorporates a narrative arc with relative layout hints (e.g., ``place B below A'') rather than absolute coordinates, granting the system flexibility while preserving structural intent. Subsequently, the storyboard is translated into executable code through the animation operator:
    \begin{equation}
        c_{anim,i} = \mathcal{F}_{\text{anim}}(o_i, \mathcal{R}_{\text{spec}})
    \end{equation}
    The operator $\mathcal{F}_{\text{anim}}$ injects aesthetic presets (color palettes, text sizes) from $\mathcal{R}_{\text{spec}}$ while optimizing object coordinates within recommended safe regions. Finally, the generated code is processed through $\textsc{Exec}(c_{anim,i})$ utilizing \texttt{python-manim} to render the animation clip $\mathcal{V}$ in \texttt{.mp4} format, with the synchronized narration script produced as $a_i = \mathcal{F}_{\text{narr}}(o_i, \mathcal{R}_{\text{spec}})$.
    
\end{itemize}

\begin{figure*}[htbp]
     \centering
     \begin{subfigure}[b]{0.35\textwidth}
         \centering
         \includegraphics[height=3.3cm]{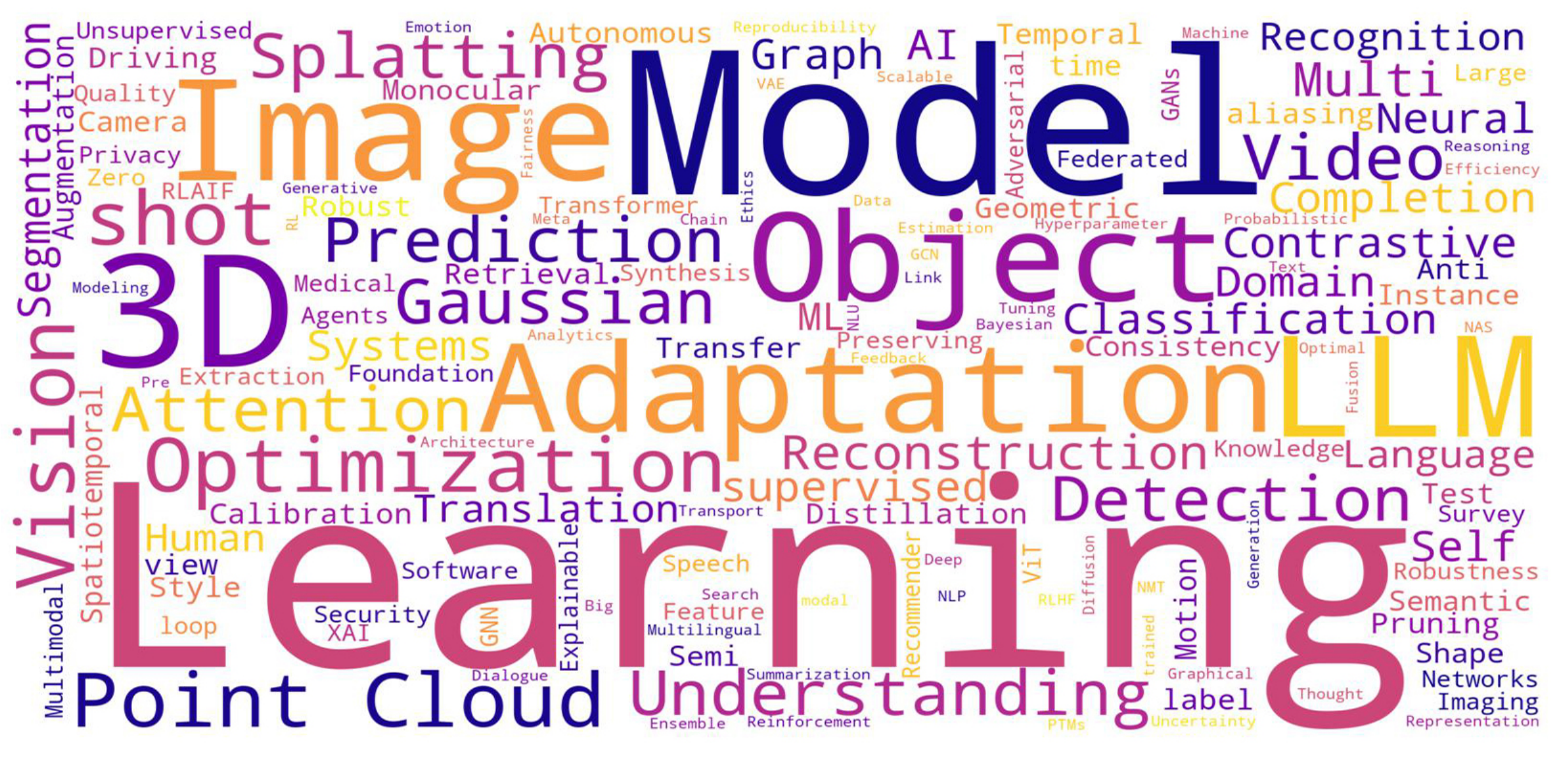}
         \caption{Keywords distribution}
         \label{fig:keywords}
     \end{subfigure}
     \hspace{5pt}
     \begin{subfigure}[b]{0.24\textwidth}
         \centering
         \includegraphics[height=3.3cm]{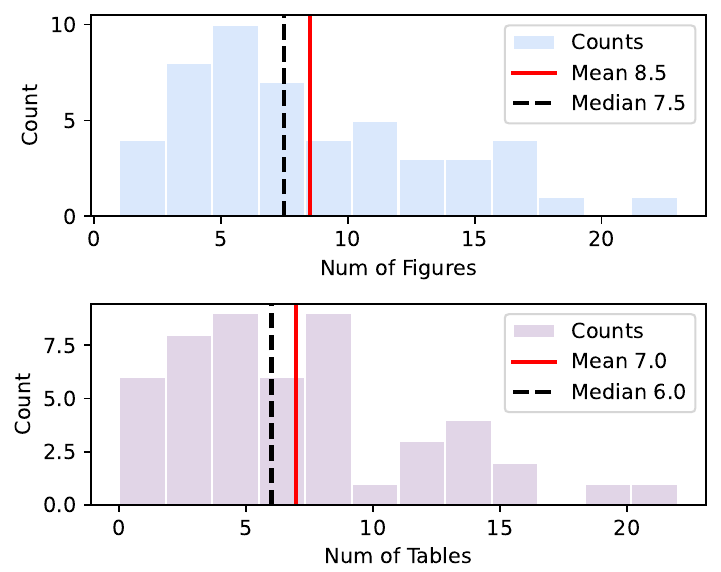}
         \caption{Paper asset stats}
         \label{fig:paper_stats}
     \end{subfigure}
     \begin{subfigure}[b]{0.35\textwidth}
         \centering
         \includegraphics[height=3.3cm]{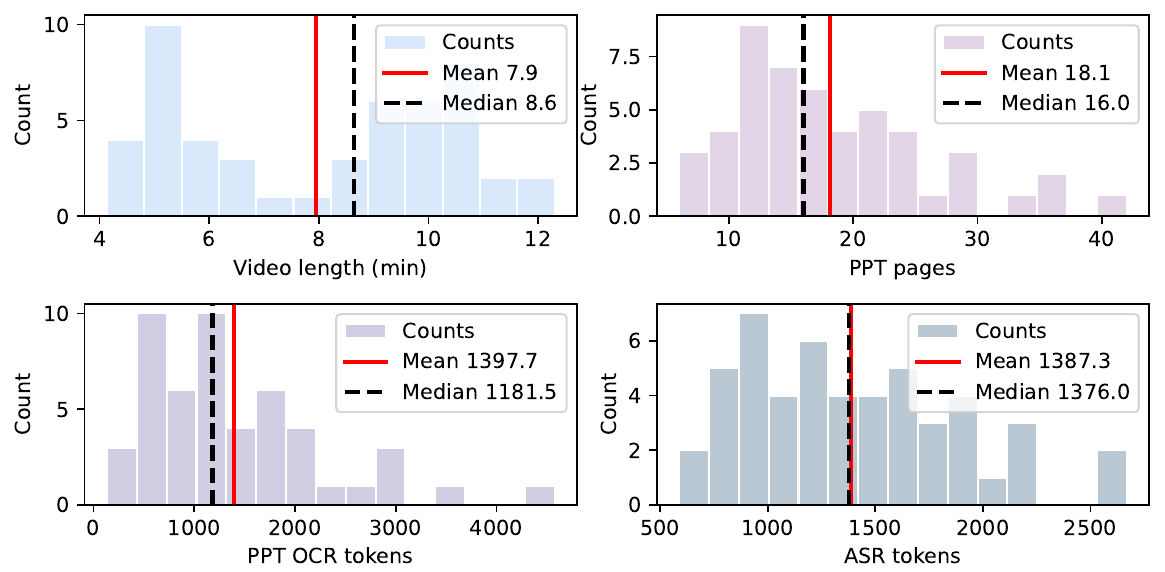}
         \caption{Video and PPT stats}
         \label{fig:video_stats}
     \end{subfigure}
     \vspace{-0.2cm}
     \caption{Statistics of our dataset. (a) word cloud of paper keywords; (b) distribution of figures and tables in papers; (c) distribution of video length, PPT pages, and tokens.}
     \label{fig:data_dist}
\end{figure*}

Narration scripts $a_i$ from all pages are aggregated into $\mathcal{A}$ for audio synthesis. Should any implementation code fail during the execution phase, the planner initiates a self-correction loop to regenerate the problematic segments. To ensure structural and visual fidelity, a VLM operator $\mathcal{F}_{\text{verify}}$ is employed to scrutinize the rendered outputs. This involves converting \texttt{.pptx} documents and uniformly sampled \texttt{.mp4} clips into \texttt{.jpg} frames to detect layout anomalies such as overlapping elements or out-of-bounds text.

\vspace{-6pt}

\subsection{Multimodal Synthesizer}
\label{sec:synthesizer}

Once the multimodal assets—comprising the audio narration set $\mathcal{A}$, the static presentation $\mathcal{P}$, and the procedural animation clips $\mathcal{V}$—are finalized, the remaining challenge lies in integrating them into a temporally coherent video. To ensure precise synchronization, the synthesizer maintains a strict mapping where the capacity of $\mathcal{A}$ corresponds one-to-one with the number of pages in $\mathcal{P}$, while each clip in $\mathcal{V}$ is aligned with its respective placeholder slide. The synthesis process begins by converting each narration script $a_i \in \mathcal{A}$ into an audio clip via a \texttt{text-to-speech} (TTS) engine \cite{radford2023robustwhisper}. The duration of each resulting audio clip serves as the primary temporal reference for its corresponding narrative node. Specifically, the playback speed of the visual modalities is adaptively adjusted to match the audio flow: static slide pages in $\mathcal{P}$ are displayed for the exact length of their narration, and dynamic animations in $\mathcal{V}$ are time-scaled to align with the explanatory pace. Finally, \texttt{MoviePy} \cite{moviepy} is employed to composite the time-aligned visual sequences, narration audio, and subtitles into the final \texttt{.mp4} video, utilizing the resolution and frame rate parameters defined in $\mathcal{R}_{\text{spec}}$ to ensure a seamless multimodal presentation.

\vspace{-10pt}
\section{Benchmark and Evaluation}
\label{sec:benchmark}

To evaluate the quality of personalized scientific videos synthesized from papers, we encounter two primary challenges: the scarcity of high-quality paper-video datasets and the technical difficulty of extracting granular semantic information for objective comparison. To address these, we present \textbf{\textit{SciVidEval}}, a comprehensive benchmark comprising a curated paper-video dataset, an automated processing toolchain, and a multi-dimensional metric suite. We detail our data (\S~\ref{sec:data}), extraction methodology (\S~\ref{sec:extraction}), and evaluation framework (\S~\ref{sec:metrics}) below.

\vspace{-6pt}
\subsection{Data Collection and Distribution}
\label{sec:data}

\textbf{Data Source.} We focus on AI research papers from multiple domains, collecting author-created oral presentation videos from top-tier conferences over the last three years, including CVPR, ICML, ICLR, ACL, and NeurIPS. These videos accompany peer-reviewed papers and represent the authors' distillation of their work, providing a natural reference standard for scientific communication.

\noindent \textbf{Data Distribution.} Our benchmark comprises 50 high-quality paper-video pairs across critical domains: Computer Vision (26\%), NLP (22\%), Multimodality (12\%), and Machine Learning (40\%). The inherent complexity of these papers—averaging 15.6 paper pages and 15.5 visual assets (see Fig.~\ref{fig:paper_stats})—poses a significant cross-modal alignment challenge. Successfully transforming this dense information into concise 4–12 minute videos requires sophisticated content selection that preserves structural integrity. The wide variance in slide counts and token densities (see Fig.~\ref{fig:video_stats}) further underscores the necessity for an intelligent system like VideoAgent, which can move beyond simple template-based tools to achieve content-aware summarization and adaptive synthesis.
SciVidEval should therefore be viewed as an initial benchmark centered on AI/ML papers; extending it to domains with less algorithmic visuals may require different rendering backends.

\begin{figure*}[t]
    \centering
    \includegraphics[width=\textwidth]{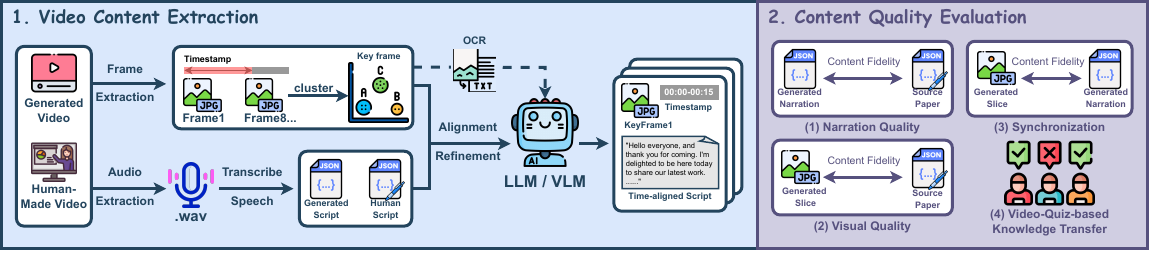}
    \vspace{-0.7cm}
    \caption{The \textbf{\textit{SciVidEval}} process involves: \textbf{1. Video Content Extraction} to obtain keyframes and narration scripts, and \textbf{2. Content Quality Evaluation}, which assesses video content against the paper on narration, visual, and synchronization quality.}
    \label{fig:evaluation_framework}
\end{figure*}

\subsection{Video Content Extraction}
\label{sec:extraction}

For automated and fine-grained evaluation, we first transform the continuous multimodal streams of scientific videos into structured, time-aligned semantic segments to facilitate a direct comparison with the source paper. We develop an automated processing toolchain to construct this representation from raw videos. Given a presentation video, we decompose it into an audio stream $\mathcal{A}$ and a visual frame sequence $\mathcal{V} = \{v_i\}_{i=1}^n$ sampled at a fixed interval $\Delta \tau$, and process them as follows:

\textbf{1) Frame Extraction.} Scientific presentation videos are typically characterized by long intervals of static visual content, such as slides or figures, with infrequent transitions. To avoid redundant processing and significantly reduce the computational overhead of optical character recognition (OCR), we first partition the visual sequence into stable segments representing distinct presentation states. We extract feature embeddings $\mathcal{E} = \{\phi(v_i)\}_{i=1}^n$ using the vision encoder of AltCLIP \cite{AltCLIP}, which aligns visual features with long-form text to ensure cross-modal semantic consistency. We then apply DBSCAN \cite{ester1996densityDBScan} clustering on $\mathcal{E}$ (with eps 0.15 and min samples 2) to group visually similar frames into clusters. Each cluster represents a continuous visual segment, from which the terminal frame is selected as the keyframe $k$. OCR is then applied to each keyframe using DeepSeekOCR \cite{deepseekocr} (or lightweight PaddleOCR \cite{PaddleOCR}) to extract the slide text $T_{ocr}$. Consequently, each visual segment is formalized as $\mathcal{S}_{\mathcal{V}} = (T_{ocr}, k, \tau_{\mathcal{V}}^{start}, \tau_{\mathcal{V}}^{end})$.

\textbf{2) Audio Extraction.} The audio stream $\mathcal{A}$ is processed via an automatic speech recognition (ASR) model (e.g., Paraformer \cite{Gao2022ParaformerFA}) equipped with a pause detection mechanism. This generates a sequence of speech segments partitioned by sentence-ending pauses. Each segment is formalized as an acoustic descriptor $\mathcal{S}_{\mathcal{A}} = (T_{asr}, \tau_{\mathcal{A}}^{start}, \tau_{\mathcal{A}}^{end})$, where $T_{asr}$ represents the transcribed text.
    
\textbf{3) Alignment and Refinement.} To ensure the narration is evaluated within the correct visual context, we align speech segments with visual intervals. Since speech segments typically outnumber visual segments, we implement a containment-based strategy to aggregate consecutive segments $\{\mathcal{S}_{j, \mathcal{A}}, \dots, \mathcal{S}_{j+n, \mathcal{A}}\}$ that fall within the duration of a visual segment $\mathcal{S}_{i, \mathcal{V}}$. This alignment is followed by a LLM-based refinement to correct potential OCR and ASR errors. This process yields a sequence of semantic segments $\mathcal{S}$, where each chapter pairs slide text $T_{ocr}$ with its corresponding spoken narration $T_{asr}$, transforming raw video into a structured series of knowledge units ready for joint multimodal evaluation.

\subsection{Evaluation Metrics}
\label{sec:metrics}
Based on the aligned visual text $T_{ocr}$ and narration $T_{asr}$, \textbf{\textit{SciVidEval}} evaluates the quality of synthesized scientific videos across two dimensions: \textbf{Content Quality} and \textbf{Knowledge Transfer}. 

\noindent\textbf{Content Quality} assesses the intrinsic quality of the generated multimodal artifacts and their content consistency with the source paper through three core metrics:

\begin{itemize}[left=0pt]
     \item \textbf{Narration Quality:} To ensure linguistic fluency and technical grounding, we measure fluency via Perplexity (PPL) calculated by \texttt{GPT-2}~\cite{radford2019language} on the concatenated $T_{asr}$ from all segments. For content fidelity, we quantify the overlap between $T_{asr}$ and the refined paper content $\mathcal{D}$ using Rouge-L \cite{lin-2004-rouge}. 
     Furthermore, semantic consistency is rated on a 0–10 scale by an LLM-as-Judge as a scalable proxy, focusing on whether the narration accurately conveys the paper’s essential technical claims rather than strictly adhering to the original phrasing.

    \item \textbf{Visual Quality:} This evaluates slide layout and technical fidelity. Slide text fluency is measured using PPL on $T_{ocr}$ extracted via OCR, ensuring the presentation is readable. To verify the contextual relevance of visual assets, image-text correspondence is a binary metric ($0$ or $1$) judged by GPT-4o, checking if the figures or tables visually support the specific assertions on each slide. Additionally, a semantic fidelity score ($0$--$10$) assesses the accuracy of the synthesized visual content in representing the original research findings.
    
    \item \textbf{Synchronization:} To measure audio-visual temporal alignment, AltCLIP~\cite{AltCLIP} is utilized to compute embedding similarity (clip score in $[0, 1]$) between frame segments and transcripts $T_{asr}$. This is complemented by a GPT-4o logic consistency score ($0$--$10$) that ensures visual contents are synchronized with the technical concepts referenced in the narration, verifying that the multimodal stream functions as a cohesive narrative.

\end{itemize}


\noindent\textbf{Knowledge Transfer Evaluation.} To ensure viewers can comprehend the paper content, we design a Q\&A-based framework covering four core categories: \textit{Motivation}, \textit{Methodology}, \textit{Experimental Results}, and \textit{Key Contributions}. \textbf{Automated VLM Scoring.} This metric tasks GPT-4o with answering four multiple-choice questions (MCQs) per paper using the full video stream (frames + transcripts) as input, yielding an Accuracy (\%) that measures how effectively the video conveys core scientific content to an automated agent. \textbf{Human Evaluation.} A blind study was conducted involving 16 graduate students. Participants were tasked with answering the same set of MCQs after consuming the video content, providing a measure of Human Accuracy to validate the pedagogical utility of the generated content in real-world academic communication.

\begin{table*}[htbp]
  \centering
    \caption{Fundamental information extraction and content quality evaluation of \textit{\textbf{VideoAgent}} against diverse baselines.}
    \vspace{-0.3cm}
    \resizebox{0.9\textwidth}{!}{%
    \begin{tabular}{l ccc ccc cc cc}
    \toprule
    \textbf{Method} & \multicolumn{3}{c}{\textbf{Narration Quality}} & \multicolumn{3}{c}{\textbf{Visual Quality}} & \multicolumn{2}{c}{\textbf{Synchronization}} & \multicolumn{2}{c}{\textbf{Video-Quiz (\%) }} \\
\cmidrule(lr){2-4}\cmidrule(lr){5-7}\cmidrule(lr){8-9}\cmidrule(lr){10-11}
          & PPL ↓ & Rouge-L ↑ & LLM-as-Judge ↑ & PPL ↓ & Asset Match Acc. (\%) ↑ & VLM-as-Judge ↑ & CLIP-Score ↑ & VLM-as-Judge ↑ & VLM-as-Judge↑ & Human Eval. \\
    \midrule
    \multicolumn{11}{l}{\textit{Oracle Baseline}} \\
    Source Paper    & 17.46 & 1.00  & 10.00         & N/A   & N/A   & 10.00 & N/A   & N/A   & 99.50 & 95.00$\pm$2.50 \\
    Author          & 23.93 & 0.11  & 8.13          & 19.46 & 88.62 & 4.74  & 0.596 & 5.37  & 97.00 & 90.00$\pm$7.50 \\
    \midrule
    \multicolumn{11}{l}{\textit{Commercial Service}} \\
    LunWenShuo \cite{lunwenshuo}     & 22.76 & 0.12  & 8.60          & 24.08 & 48.57   & 3.76  & 0.615 & 5.45  & 89.50 & 79.50$\pm$5.50 \\
    Pictory \cite{pictory}           & 25.68 & 0.06  & 9.00          & N/A   & N/A   & 3.26  & \textbf{0.643} & 4.42  & 85.50 & 74.00$\pm$4.50 \\
    NotebookLM \cite{notebooklm}    & \textbf{16.67} & 0.07  & 8.76          & \textbf{15.37} & 31.99 & 3.16  & 0.102 & \underline{6.33}  & \underline{99.00} & 86.00$\pm$4.00 \\
    \midrule
    \multicolumn{11}{l}{\textit{Open Source Baseline}} \\
    AutoSlides \cite{Yang2025AutoSlidesAI}     & N/A   & N/A   & N/A           & \underline{17.31} & 38.37 & 6.64  & N/A   & N/A   & 98.50 & 78.00$\pm$6.50 \\
    PresentAgent \cite{Shi2025PresentAgentMA}   & N/A   & N/A   & N/A           & 26.64 & 66.75 & 5.97  & N/A   & N/A   & 97.00 & 75.00$\pm$5.00 \\
    \midrule
    \multicolumn{11}{l}{\textit{VideoAgent Variants with Default Sequential Summary}} \\
    GPT-4o \cite{hurst2024gpt} & \underline{18.08} & \textbf{0.16}  & \underline{9.38}  & 25.78 & \underline{73.31} & \underline{7.70}  & 0.634 & 6.02  & \underline{99.00} & - \\
    Gemini-2.5 Pro \cite{comanici2025gemini} & 18.32 & \underline{0.14}  & \textbf{9.70}  & 21.33 & \textbf{79.24} & \textbf{8.03}  & \underline{0.635} & \textbf{6.58}  & \textbf{99.50} & \textbf{87.50$\pm$5.00} \\
    Qwen-2.5VL-7B \cite{bai2025qwen2}  & 18.42 & 0.08  & 9.31  & 27.92 & 63.65 & 7.37  & 0.303 & 2.87  & 98.00 & - \\
    \bottomrule  
    \end{tabular}%
    }
  \label{tab:main_results}%
\end{table*}

\section{Experiments}
\label{sec:experiment}

\subsection{Baselines and Settings}
\label{sec:baselines}

To evaluate extraction and content quality, we compare \textbf{\textit{VideoAgent}} against four baseline categories: 1) \textbf{Oracle Baselines}, where the source paper and author-created video serve as upper bounds; 2) \textbf{Commercial Services}: LunWenShuo~\cite{lunwenshuo}, Pictory~\cite{pictory}, and NotebookLM~\cite{notebooklm}; 3) \textbf{Open Source Baselines}: AutoSlides~\cite{Yang2025AutoSlidesAI} and PresentAgent~\cite{Shi2025PresentAgentMA}; and 4) \textbf{VideoAgent Variants} based on GPT-4o~\cite{hurst2024gpt}, Gemini-2.5 Pro~\cite{comanici2025gemini}, and Qwen-2.5VL-7B~\cite{bai2025qwen2}.


\subsection{Quantitative Comparison}
\label{sec:results}

\noindent \textbf{Overall Performance.} As shown in Table~\ref{tab:main_results}, \textit{\textbf{VideoAgent}} variants using sequential summary consistently outperform existing baselines. Regarding narration, open-source baselines focus solely on automated PPT generation and lack integrated narration. NotebookLM produces highly natural, conversational scripts with a superior PPL of 16.67, yet VideoAgent achieves the highest content consistency according to Rouge-L and LLM-as-Judge scores. Regarding visuals, Pictory's text-to-realistic generation fails completely to match technical paper assets. LunWenShuo's template-based PPT approach with compressed scripts suffers from a higher visual PPL of 24.08. Although both LunWenShuo and NotebookLM extract certain paper assets, frequent omissions compromise their consistency, resulting in match accuracies of only 48.57\% and 31.99\%, respectively. Overall, \textit{\textbf{VideoAgent}} (Gemini 2.5 Pro) demonstrates strong competitiveness across all dimensions, achieving a VLM-judged visual quality of 8.03 and a synchronization score of 6.58.

\noindent\textbf{Knowledge Transfer.} Automated VLM-as-Judge scores benchmark knowledge transfer. \textit{\textbf{VideoAgent}} (Gemini 2.5 Pro) achieves 99.50\%. NotebookLM also performs remarkably well with a score of 99.00\%. For open-source baselines, AutoSlides achieves 98.50\% due to its higher text density, while PresentAgent reaches 97.00\% with more concise descriptions. In parallel, Pictory is hindered by irrelevant visuals, and LunWenShuo is limited by its rigid templates. Human evaluation results align with these findings; \textit{\textbf{VideoAgent}} reaches 87.5\% accuracy, approaching the level of author-created videos (90.0\%) and confirming its practical utility for academic communication.

\subsection{Further Analysis}
\label{sec:analysis}

While the previous section established \textbf{\textit{VideoAgent}}'s extraction effectiveness, this section investigates its comprehensive capabilities through three key research questions (RQs).

\captionsetup[subfigure]{font=footnotesize, justification=centering}
\begin{figure}[htbp]
  \centering
  \hspace{-0.75cm}
  \begin{subfigure}{0.6\columnwidth}
    \centering
    \includegraphics[height=3.2cm]{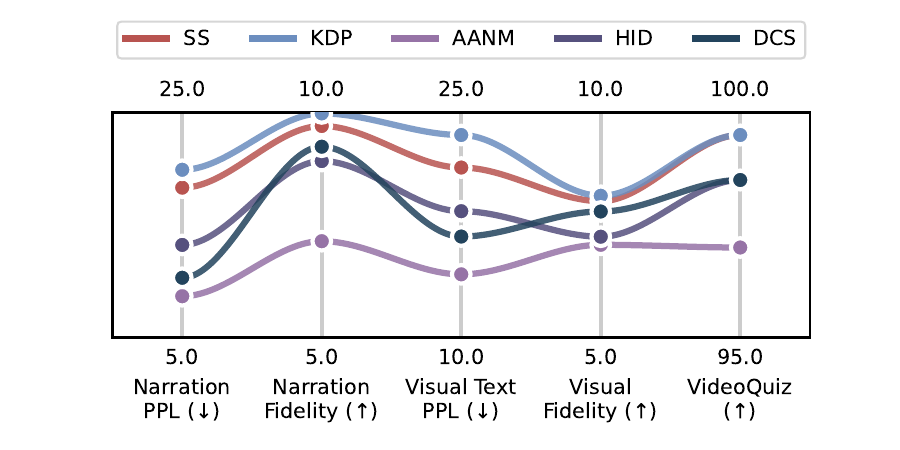}
    \vspace{-0.5cm}
    \caption{Content Quality Evaluation}
    \label{fig:parallel}
  \end{subfigure}
  \hfill
  \begin{subfigure}{0.38\columnwidth}
    \centering
    \includegraphics[height=3.2cm, keepaspectratio]{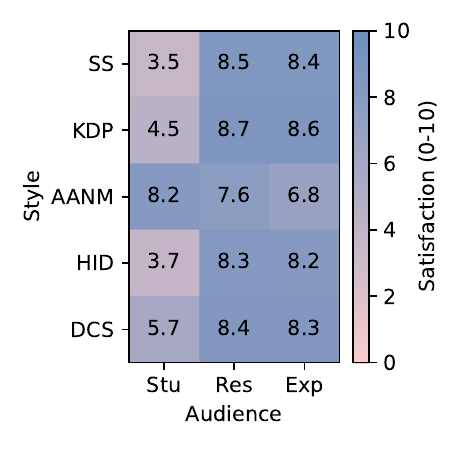}
    \vspace{-0.5cm}
    \caption{Audience Satisfaction}
    \label{fig:heatmap}
  \end{subfigure}
  \vspace{-0.3cm}
  \caption{Analysis of personalization strategies: (a) overall performance and (b) simulated audience satisfaction (0--10).}

  \label{fig:personalization_analysis}
\end{figure}

\noindent\textbf{RQ1: Personalization Effectiveness and Persona Diversity.} Beyond simple classification, we analyze stylistic divergence across 5 personas detailed in Table \ref{tab:outline_example}: \textit{Sequential Summary} (SS), \textit{Knowledge Distillation Paradigm} (KDP), \textit{Audience-Adaptive Narrative Modeling} (AANM), \textit{Heuristic Insight Discovery} (HID), and \textit{Dialectical \& Critical Synthesis} (DCS). As shown in Figure \ref{fig:parallel}, while VideoQuiz scores exhibit a ceiling effect ($>95\%$ accuracy), distinct strategies prioritize different quality facets. For instance, AANM prioritizes accessibility with the lowest PPL in both narration and visual text, while KDP maintains higher fidelity at the cost of linguistic complexity. This divergence translates into varying target audience alignment, as visualized in Figure \ref{fig:heatmap}. Notably, AANM achieves a peak satisfaction score of 8.2 for Students (Stu), outperforming more technical personas like DCS (5.7). Conversely, for professional Researchers (Res) and Experts (Exp), the more structured KDP is preferred, scoring 8.7 and 8.6, respectively. These findings confirm that \textbf{\textit{VideoAgent}} does not merely follow instructions but substantially restructures the technical narrative to meet specific user needs.

\begin{table}[t]
\centering
\caption{Evaluation of Manim animation synthesis quality.}
\label{tab:manim}
\vspace{-0.3cm}
\resizebox{0.9\columnwidth}{!}{
\begin{tabular}{llccc} 
\toprule
\textbf{Method} & \textbf{Backbone} & \textbf{Pass@1/3}$\uparrow$ & \textbf{Logic}$\uparrow$ & \textbf{Consist.}$\uparrow$ \\
\midrule
\multirow{2}{*}{Manimator}  
  & Gemini-2.5 Pro    & 58\%/84\%  & 4.62 & 4.23 \\
  & Claude-4.5-Opus   & 82\%/100\% & 4.60 & 4.30 \\
\midrule
\multirow{2}{*}{Ours}  
  & Gemini-2.5 Pro    & 78\%/100\% & 5.00 & 4.86 \\
  & Claude-4.5-Opus   & 90\%/100\% & 4.90 & 4.91 \\
\bottomrule
\end{tabular}%
} 
\vspace{1mm}
\footnotesize
\\ \textit{Note:} For Manimator, Pass@3 indicates the success rate of the best result among 3 independent generations. For our method, Pass@3 reports success after applying self-correction (max 3 turns).
\end{table}

\begin{table}[htbp]
  \centering
  \caption{Resource and Token Efficiency Analysis.}
  \label{tab:system_robustness}
  \vspace{-0.3cm}
  \resizebox{0.9\columnwidth}{!}{%
    \begin{tabular}{llr}
    \toprule
    \textbf{Module} & \textbf{Function} & \textbf{Avg. Tokens ($T_{in}$ / $T_{out}$)} \\
    \midrule
    \multirow{2}{*}{\textbf{Document Parser}} & Paper Parsing & 9.9k / 0.9k \\
          & Asset Filtering & 2.8k / 1.1k \\
    \midrule
    \textbf{Requirement Analyzer} & Intent Parsing & 0.5k / 0.2k \\
    \midrule
    \textbf{Personalized Planner} & Outline Gen. & 2.8k / 1.5k \\
    \midrule
    \multirow{3}{*}{\textbf{Static Synthesis}} & Narration Gen. (8 pages) & 1.2k / 0.6k \\
          & Bullet Gen. (8 pages) & 2.0k / 1.4k \\
          & Image Gen. (5 pages) & 1.2k / 13.2k \\
    \midrule
    \textbf{Dynamic Synthesis} & Manim Gen. (1 clip) & 12.2k / 11.8k \\
    \midrule
    \multicolumn{2}{r}{\textbf{Total Processed:}} & $T_{in}$: 32.6k \quad $T_{out}$: 30.7k \\
    \bottomrule
    \end{tabular}%
  }
  \vspace{1mm}
\footnotesize
\\ \textit{Note:} All token counts represent the average total consumption per video across the indicated number of pages/clips.
\end{table}

\begin{figure*}[t]
    \centering
    \includegraphics[width=0.87\textwidth]{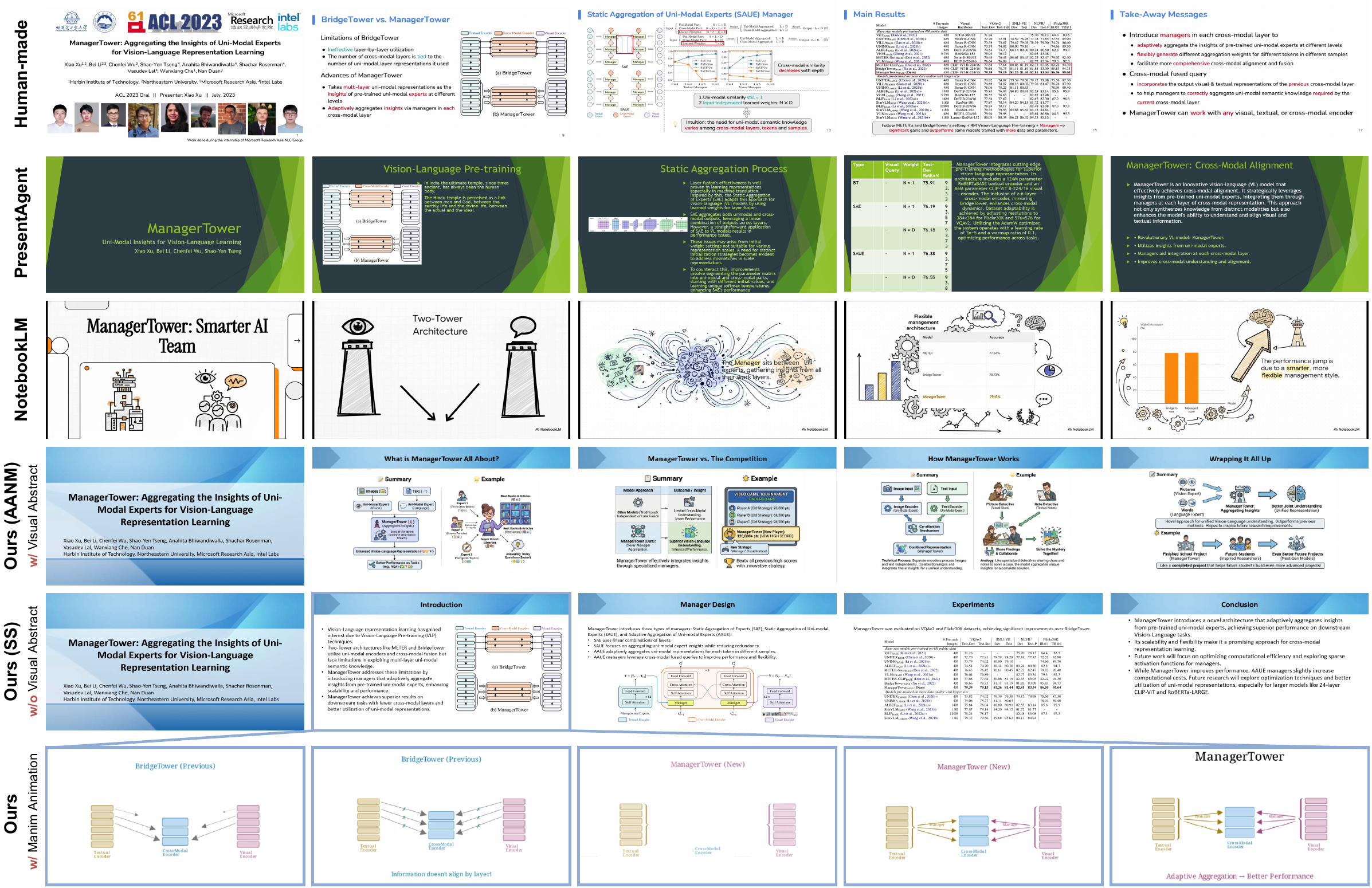}
    \vspace{-0.3cm}
    \caption{Case study of \textbf{\textit{VideoAgent}} comparing integrated synthesis with open-source and commercial solutions.}
    \label{fig:case_study}
\end{figure*}

\noindent\textbf{RQ2: Manim Code Generation Quality.}
We evaluate our animation synthesis module against Manimator~\cite{P2025Manimator}. Inspired by TheoremExplainBench~\cite{Ku2025TheoremExplainAgent}, we report three metrics aligned with our design goals: \textit{Pass Rate} measures whether generated code complies and renders successfully; \textit{Logical Flow} assesses whether the animation presents concepts in a coherent narrative progression; \textit{Visual Consistency} evaluates whether the motions are smooth, and whether the visual style remains uniform across frames. The latter two metrics are rated out of 5. As shown in Table~\ref{tab:manim}, our method achieves higher Pass@1 rates across backbones (90\% vs.\ 82\% with Claude-4.5-Opus, 78\% vs.\ 58\% with Gemini-2.5 Pro). While both methods reach near-perfect Pass@3, they differ in mechanism: Manimator samples multiple independent generations and selects the best, whereas ours applies self-correction on a single generation. For content quality, we observe consistent gains in Logical Flow (5.00 vs.\ 4.62 with Gemini-2.5 Pro), suggesting that the structured storyboard with pre-specified narrative arcs helps maintain explanatory coherence. The improvement in Visual Consistency (4.91 vs.\ 4.30 with Claude-4.5-Opus) indicates that style injection and visual validation via $\mathcal{F}_{\text{verify}}$ effectively enforce uniform aesthetics. 


\noindent\textbf{RQ3: Resource and Token Efficiency Analysis.} To evaluate \textbf{\textit{VideoAgent}}'s computational cost, we analyze token usage and monetary expense across 50 papers via Gemini 2.5 Pro, as end-to-end generation time varies by service and hardware.
Each generated video averages 8 pages (spanning Title to Conclusion), with 5 pages requiring image-generation (\texttt{Nano Banana}~\cite{google2025nanobanana}) and 1 page requiring animation generation (\texttt{Manim}). The document parser handles the highest input load (12.7k tokens) for semantic extraction. The personalized planner structures the narrative, while the static and dynamic synthesis modules drive the output volume. Notably, Manim generation demands the highest bidirectional throughput (12.2k input / 11.8k output) to ensure precise code synthesis. Total processing averages 32.6k input and 30.7k output tokens per video, resulting in an average cost of approximately \$0.35 per video. These results indicate that \textbf{\textit{VideoAgent}}'s resource consumption is modest, making per-paper video synthesis economically feasible at scale.

\subsection{Case Study}
As shown in Figure \ref{fig:case_study}, existing open-source methods like \textit{PresentAgent} frequently encounter layout inconsistencies and overflow issues. While commercial services such as \textit{NotebookLM} provide polished visual effects, they often remain too abstract for precise technical communication. In contrast, \textbf{\textit{VideoAgent}} (Rows 4--6) achieves high-fidelity visual quality comparable to human-made standards. Specifically, Row 4 (AANM) provides vivid narratives tailored for junior audiences; Row 5 features adaptive layouts for automated content synthesis, and Row 6 utilizes \texttt{Manim} for precise animated explanations of visual assets. This framework strikes a pragmatic balance between automation and grounded fidelity, offering a reliable methodology for representing complex research insights.

\section{Conclusions}
\label{sec:conc}
In this work, we present \textbf{\textit{VideoAgent}}, an intent-driven and modular framework that formulates scientific video generation as a personalized planning problem. VideoAgent enables the coherent integration of static slides, procedural animations, and synchronized narration through a staged pipeline with self-correction loop. We further introduce \textbf{\textit{SciVidEval}}, a task-oriented benchmark that evaluates scientific videos by combining automated multimodal quality metrics with a Video-Quiz-based protocol to assess communicative effectiveness. Experiments on 50 real paper-video pairs show that VideoAgent achieves competitive content quality and knowledge transfer compared to both commercial services and author-created videos. Nevertheless, the quality of dynamic animations remains constrained by the underlying generative models, highlighting the need for more robust metrics and verification signals for procedural animation correctness and temporal grounding.

\begin{acks}
This work was supported by the National Natural Science Foundation of China (62577041, 62192730, 62192734), the Fundamental and Interdisciplinary Disciplines Breakthrough Plan of the Ministry of Education of China (JYB2025XDXM104), the Outstanding Youth Science Foundation of Shaanxi Province (2025JC-JCQN-083), the Key Research and Development Program of Shaanxi Province (2025CY-YBXM-047), the CCF-Huawei Populus Grove Fund (CCFHuawei-FM202507), and the Natural Science Foundation of Xi'an (2025JH-ZRKX-0540).
\end{acks}

\clearpage

\bibliographystyle{ACM-Reference-Format}
\bibliography{refs}

\end{document}